\documentclass[11pt]{article}

\usepackage[preprint]{acl}

\usepackage{times}
\usepackage{latexsym}

\usepackage[T1]{fontenc}

\usepackage[utf8]{inputenc}

\usepackage{microtype}

\usepackage{inconsolata}

\usepackage{graphicx}

\usepackage{booktabs}

\title{SkillWiki:
A Living Knowledge Infrastructure for Agent Skills}

\author{
Dingcheng Huang$^{1}$,
Yuda Ding$^{1}$,
Bingshuo Liu$^{1}$,
Qingbin Liu$^{2}$,
Xi Chen$^{2}$, Jiang Bian$^{2}$, \\
\textbf{Hongliang Sun$^{1}$}\textbf{,}
\textbf{Zhiying Tu$^{1}$}\textbf{,}
\textbf{Dianhui Chu$^{1}$}\textbf{,}
\textbf{Xiaoyan Yu}$^{3}$\textbf{,}
\textbf{Dianbo Sui}$^{1}$\thanks{Corresponding author.}\\
$^{1}$Harbin Institute of Technology\\
$^{2}$Tencent  \\
$^{3}$Nanyang Technological University  \\
\texttt{dingcheng@stu.hit.edu.cn}, \quad \texttt{suidianbo@hit.edu.cn}
}

\begin{document}
\maketitle
\begin{abstract}
While knowledge is managed through Wikipedia and software through GitHub, agent skills still lack an infrastructure for large-scale production, governance, and evolution. \textbf{SkillWiki} is a living knowledge infrastructure that supports the organization, grounding, and continuous evolution of agent skills by transforming heterogeneous knowledge into reusable skill assets linked to their originating evidence. Our demonstration presents the complete skill lifecycle, from knowledge ingestion and skill production to provenance-aware exploration, governance, and execution-driven evolution. SkillWiki highlights a future in which knowledge, skills, and execution experience co-evolve within a shared infrastructure. The live demonstration and source code are publicly available at \url{https://github.com/Huangdingcheng/SkillWiki}.
\end{abstract}



\section{Introduction}

Large language model (LLM)-based agents are rapidly evolving from systems that perform isolated tool calls into autonomous systems capable of executing complex long-horizon tasks~\cite{yao2023react, Wang2023VoyagerAO, Liu2023AgentBenchEL, Xie2024OSWorldBM, Wang2024OpenHandsAO}. To operate effectively in real-world environments, such agents must acquire a diverse set of reusable skills~\cite{Xu2026AgentSF}, ranging from tool use and interface manipulation to workflow execution, multi-agent coordination, and domain-specific problem solving. At the same time, the sources from which skills are acquired continue to expand~\cite{Wang2023VoyagerAO, wang2026webxskillskilllearningautonomous}. Beyond manually designed prompts and tool specifications, agents can increasingly learn new skills from execution trajectories, environmental interactions, documentation, and accumulated experience. As the capabilities of agents continue to grow, so does the scale of their skill collections, leading to increasingly large and heterogeneous repositories of reusable skills.

\begin{figure}[t]
    \centering
    \includegraphics[width=1\columnwidth]{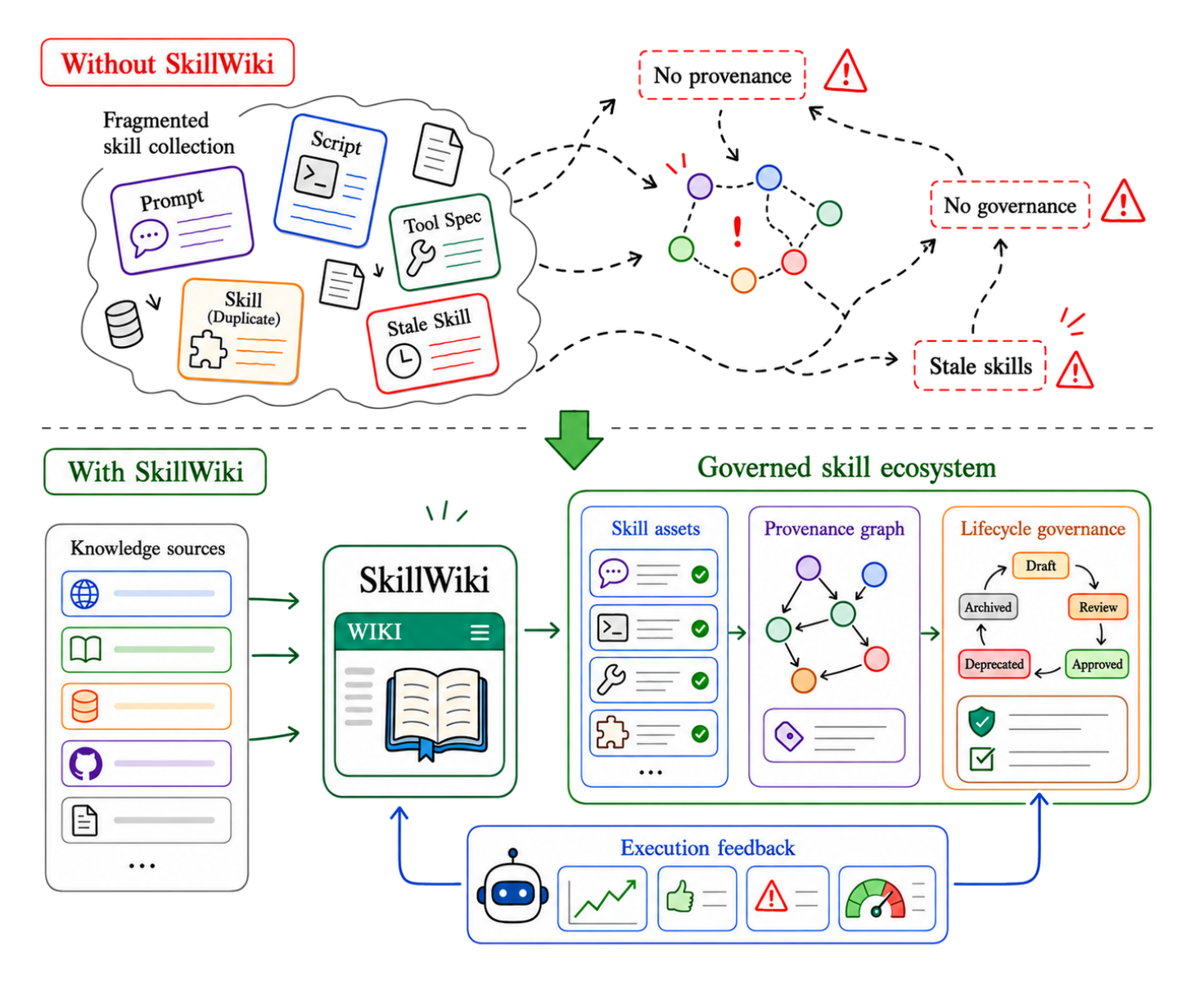} 
    \caption{
    From skill collections to governed skill ecosystems.
    }
    \label{fig:intro_motivation}
\end{figure}

To cope with the growing demand for reusable skills, recent research has explored a wide range of directions, including skill acquisition~\cite{Wang2023VoyagerAO, Si2026FromCT, yang2026autoskillexperiencedrivenlifelonglearning, wang2026skillxautomaticallyconstructingskill, Zhang2026CoEvoSkillsSA, zhang2026skillflowbenchmarkinglifelongskilldiscovery}, skill memory~\cite{Zhang2026MemSkillLA, Lin2026MUSEAutoskillSA, yang2026autoskillexperiencedrivenlifelonglearning}, skill evolution, and lifecycle management~\cite{Shen2026DynamicSL, Liu2026SkillsVoteLG, Lin2026MUSEAutoskillSA}. These efforts enable agents to continuously accumulate and refine skills from trajectories, experience, and environmental feedback. However, most existing approaches focus on a particular stage of the skill lifecycle or optimize skills as capability components for individual agents. As skill collections continue to expand and absorb new knowledge from tasks, trajectories, documents, and execution experience, increasingly complex relationships emerge among skills, tools, and their underlying knowledge sources. Consequently, the challenge is no longer limited to acquiring, retrieving, or evolving individual skills, but extends to organizing, tracing, governing, and evolving an ever-growing skill ecosystem. While knowledge management benefits from collaborative platforms such as Wikipedia and software engineering relies on versioned infrastructures such as Git and GitHub, comparable organizational and governance infrastructures for agent skills remain largely absent. As skills increasingly become a critical capability asset for agents, building a unified infrastructure for large-scale skill ecosystems is emerging as a new systems challenge.

As illustrated in Figure~\ref{fig:intro_motivation}, the limitation of existing approaches does not stem from a lack of mechanisms for skill acquisition, retrieval, or evolution, but from the absence of a unified infrastructure for managing skills at scale. Knowledge management relies on collaborative platforms such as Wikipedia~\cite{Giles2005}, while software engineering benefits from infrastructures such as Git and GitHub~\cite{10.1145/2145204.2145396}. As skills increasingly become long-lived capability assets, future agent systems require a similar infrastructure to support their production, organization, governance, and evolution.

To address this challenge, we introduce SkillWiki, an autonomous skill production and governance system for large-scale agent skill ecosystems. SkillWiki continuously transforms heterogeneous knowledge materials，including trajectories, documents, API specifications, scripts, historical skills, and agent execution experience—into reusable, executable, verifiable, and versioned skill assets. It further provides a unified framework for skill organization, provenance tracking, governance, and evolution, establishing a closed-loop ecosystem in which knowledge, skills, and execution experience continuously co-evolve.

Rather than viewing skills as static units of storage, SkillWiki places them within a continuously evolving production pipeline, where new knowledge materials give rise to new skills, skill execution generates new experience, and failures or performance degradation trigger maintenance and version evolution. Through this process, SkillWiki enables a self-sustaining ecosystem in which skills are continuously produced, governed, reused, and improved over time.

In summary, our contributions are three-fold:

\begin{itemize}
    \item We introduce SkillWiki, a unified infrastructure for skill production and governance in large-scale agent skill ecosystems. Unlike prior approaches that treat skills as isolated capability units or static memory objects, SkillWiki models skills as governed capability assets that can be continuously produced, organized, maintained, and evolved, thereby reframing skill management as an infrastructure problem for agent ecosystems.
    \item We present a continuous skill production pipeline that transforms heterogeneous knowledge materials into governed skill assets. The pipeline continuously extracts and constructs skills from trajectories, documents, API specifications, scripts, historical skills, and agent execution experience, and converts them into reusable, executable, verifiable, and maintainable assets through auditing, validation, provenance tracking, version control, and evolution mechanisms.
    \item We implement a complete SkillWiki system and provide an end-to-end interactive demonstration covering knowledge ingestion, skill construction, provenance exploration, lifecycle governance, version evolution, and agent execution feedback. The system showcases a closed-loop workflow in which skills are continuously produced, used, maintained, and regenerated from heterogeneous knowledge materials.
\end{itemize}

\section{SkillWiki}
\subsection{Overview}

\begin{figure*}[!t]
    \centering
    \includegraphics[width=2\columnwidth]{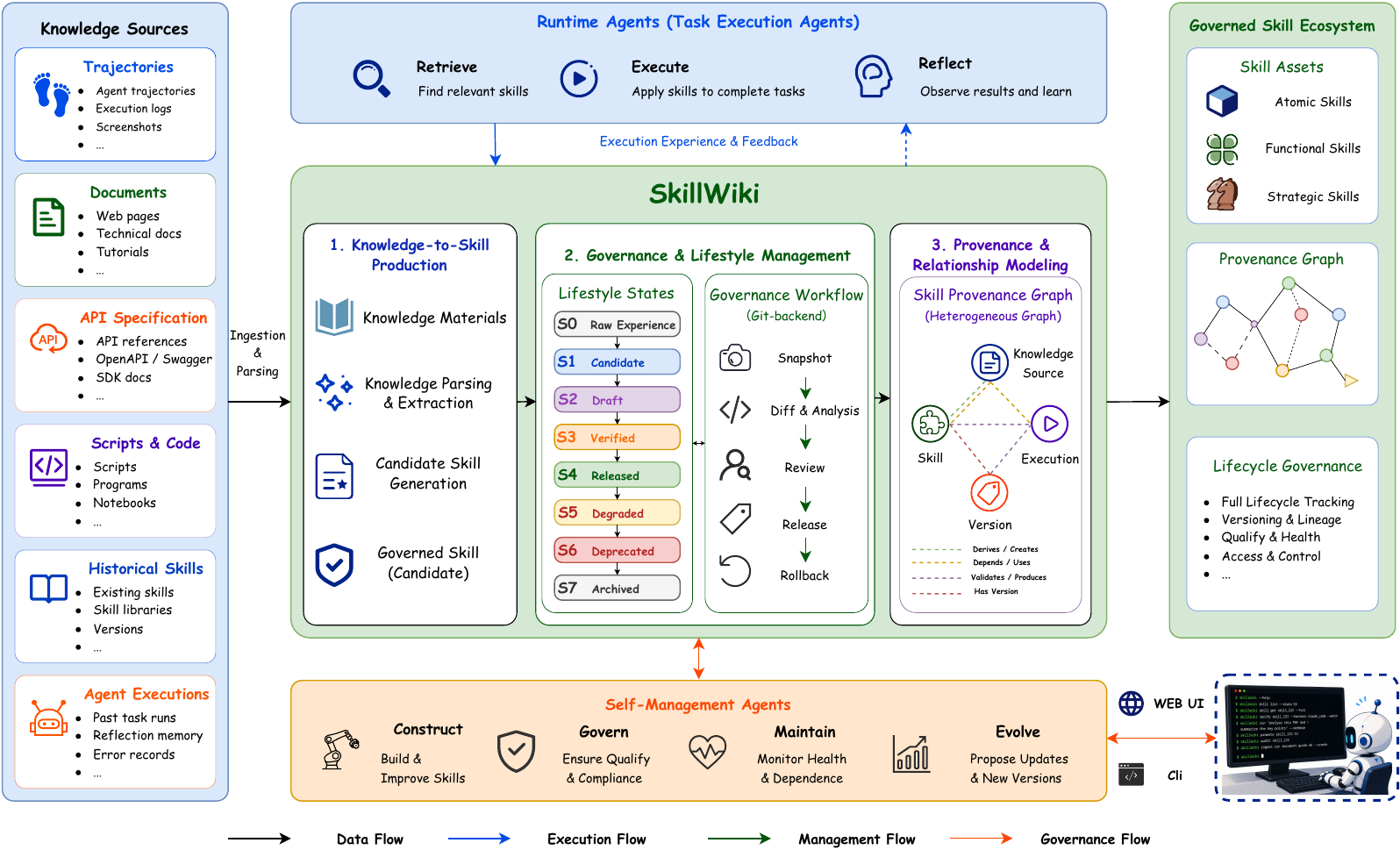}
    \caption{Overview of SkillWiki.}
    \label{fig:overview}
\end{figure*}

\textbf{SkillWiki} is a unified infrastructure for large-scale agent skill ecosystems. Inspired by the collaborative management paradigm of Wikipedia, it treats skills as governed assets rather than static memory units or repository entries. As illustrated in Figure~\ref{fig:overview}, SkillWiki consists of two tightly coupled workflows: a knowledge-grounded production workflow that transforms heterogeneous knowledge materials into governed skill assets and constructs a provenance graph linking knowledge, skills, tools, and execution evidence; and a governance workflow that supports skill organization, validation, versioning, maintenance, and evolution. Together, these workflows form a closed-loop ecosystem in which knowledge, skills, and experience continuously co-evolve.

\subsection{From Knowledge to Skills}

SkillWiki transforms heterogeneous knowledge materials into reusable skill assets. Unlike conventional skill repositories, it separates the knowledge layer from the skill layer: original knowledge materials are preserved as long-term assets, while skills are continuously constructed and evolved as executable capabilities. This process is supported by a unified pipeline for knowledge acquisition, skill construction, and relationship modeling, forming the foundation of a traceable and evolvable skill ecosystem.

\paragraph{Knowledge Sources.}
SkillWiki adopts a decoupled design in which knowledge repositories and skill repositories are maintained separately. Rather than modifying or replacing original knowledge, the system preserves it as long-term knowledge assets and continuously derives new skills from it. The knowledge layer integrates heterogeneous materials, including trajectories, documents, API specifications, scripts, historical skills, and agent execution records, which serve as both the foundation for skill construction and the evidence basis for subsequent validation, maintenance, and evolution.

\paragraph{Knowledge-Grounded Skill Construction.}
SkillWiki adopts a knowledge-grounded skill construction process that transforms heterogeneous knowledge materials into structured skill assets. Rather than directly converting documents, trajectories, or execution records into prompts, the system extracts reusable actions, workflows, and operational patterns from raw knowledge sources and organizes them into structured capability representations. Throughout this process, \textbf{the original knowledge materials are preserved as evidence sources and remain explicitly linked to the resulting skills.}

In SkillWiki, a skill is modeled as a lifecycle-aware capability asset rather than a prompt template or tool wrapper. Each skill encapsulates its identity, classification, lifecycle state, interface specification, implementation, evaluation contract, provenance information, graph relations, and runtime metrics. This unified representation enables skills to be validated, versioned, governed, composed, and continuously evolved while preserving traceable connections to their underlying knowledge evidence.

\paragraph{Skill Provenance Graph.}
Beyond individual skills, SkillWiki maintains a Skill Provenance Graph that captures lineage, dependency, and evolution relationships across the skill ecosystem. The graph connects knowledge sources, skills, executions, validations, versions, and inter-skill dependencies through a unified heterogeneous representation. By continuously recording provenance evidence, execution history, and version lineage, it provides a relational backbone for skill tracing, governance, maintenance, and evolution, enabling both users and agents to inspect how skills are created, validated, reused, and evolved over time.

\subsection{Managing Skills at Scale}

As skill ecosystems continue to grow, managing skills becomes as important as constructing them. Beyond skill creation, large-scale repositories must address challenges such as organization, quality control, dependency management, version evolution, and skill degradation. To this end, SkillWiki provides a unified framework for skill organization, governance, and evolution, while leveraging self-management agents to monitor skill health, identify maintenance needs, and generate evolution proposals, ensuring that skills remain discoverable, reliable, and maintainable over time.

\paragraph{Organizing Skills.}
SkillWiki organizes skills as structured capability assets rather than loosely stored prompts or scripts. Each skill follows a unified schema that captures its interface, implementation, evaluation, provenance, and graph relations. To support large-scale management, SkillWiki organizes user-facing skills through both a hierarchical taxonomy and a lifecycle model. The taxonomy distinguishes atomic skills for basic operations, functional skills for reusable task capabilities, and strategic skills for high-level planning and coordination. Meanwhile, the lifecycle model tracks skills across Raw Experience, Candidate, Draft, Verified, Released, Degraded, Deprecated, and Archived states. Together, the unified schema, hierarchical taxonomy, and lifecycle states enable large collections of heterogeneous skills to remain searchable, composable, and manageable.

\paragraph{Governing Skills.}
Unlike conventional skill repositories that directly modify or overwrite skills, SkillWiki places all skill changes under an auditable governance workflow. Skill creation, repair, decomposition, composition, and version updates are first represented as candidate changes and then processed through a sequence of snapshots, structured diffs, reviews, and releases. These governance procedures are carried out by a set of internal meta-skills and self-management agents, which automatically identify changes to interfaces, implementations, dependencies, and evaluation contracts, while detecting potentially breaking modifications. Inspired by Git-style software governance, SkillWiki enables skill evolution to be traceable, reviewable, and reversible, preventing uncontrolled drift during long-term evolution.

\paragraph{Evolving Skills.}
SkillWiki treats skill evolution as a feedback-driven process rather than an occasional maintenance activity. During agent execution, the system continuously collects usage statistics, failure patterns, and reflection memories to assess skill health. When performance degradation, environmental shifts, or recurring failures are detected, maintenance proposals are automatically generated and routed through the governance workflow to produce updated skill versions. By connecting execution feedback, reflection memories, and governance mechanisms, SkillWiki establishes a closed-loop evolution process that enables skills to continuously adapt to new knowledge and changing environments.

\section{Demonstration}

\begin{figure*}[t]
    \centering
    \includegraphics[width=2\columnwidth]{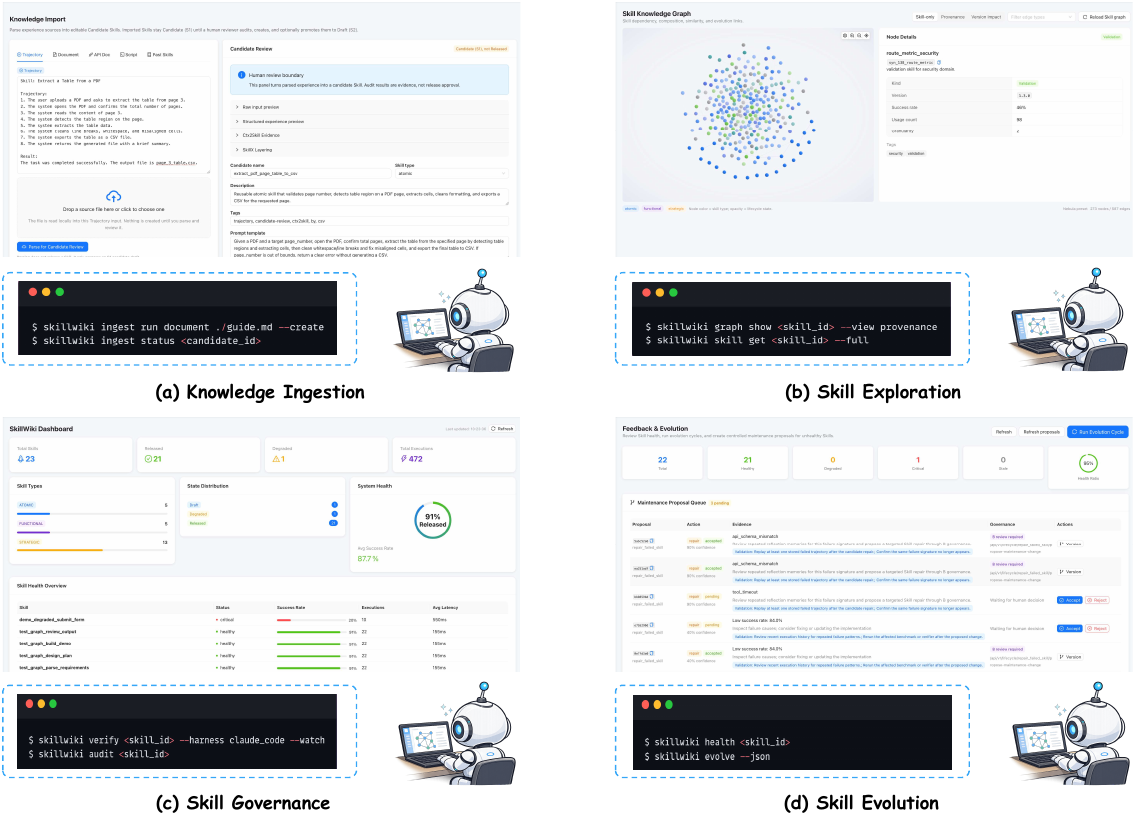}
    \caption{
    Demonstration of SkillWiki. The system supports (a) knowledge ingestion from heterogeneous knowledge materials, (b) provenance-aware skill exploration, (c) autonomous skill governance and review, and (d) execution-driven skill evolution. The same workflows can also be accessed through the SkillWiki CLI for automation and agent integration.
    }
    \label{fig:demo}
\end{figure*}

\begin{figure*}[t]
    \centering
    \includegraphics[width=2\columnwidth]{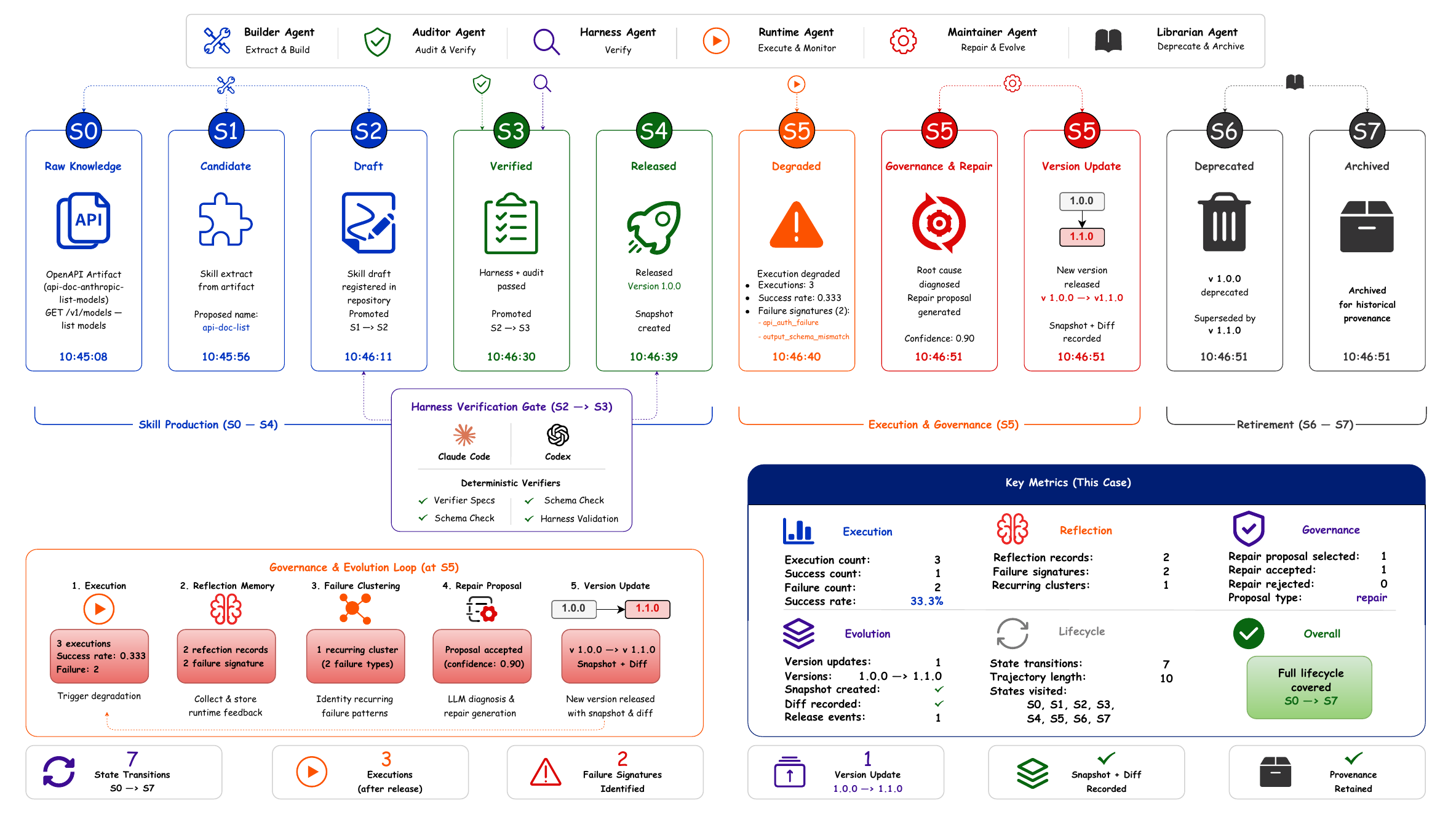}
    \caption{
    Full lifecycle trajectory of a representative skill covering all lifecycle states (S0--S7), including production, verification, release, governance-driven repair, version evolution, deprecation, and archival.
    }
    \label{fig:lifecycle}
\end{figure*}

Figure~\ref{fig:demo} presents the SkillWiki interface through four representative workflows: knowledge ingestion, skill exploration, skill governance, and skill evolution. Beyond the web interface, SkillWiki also provides a \textbf{command-line interface (CLI)} that exposes the same infrastructure for automation and agent integration. Together, these workflows demonstrate how heterogeneous knowledge materials are transformed into governed skill assets and continuously maintained through provenance tracking, governance, and execution-driven evolution.

\subsection{Knowledge Ingestion}

Figure~\ref{fig:demo}(a) illustrates the knowledge ingestion interface of SkillWiki. Users can continuously import heterogeneous knowledge materials, including trajectories, documents, API specifications, scripts, historical skills, and execution records. After ingestion, the system automatically parses the input content, extracts reusable capability patterns, and generates candidate skills together with their associated evidence and provenance information. The resulting skills are linked to their originating knowledge sources and become available for subsequent validation, governance, and evolution, establishing the first step of the skill production pipeline.

\subsection{Skill Exploration}

Figure~\ref{fig:demo}(b) presents the skill exploration interface. Beyond conventional skill retrieval, SkillWiki enables users to inspect the provenance, dependencies, execution history, and version lineage of individual skills through the Skill Provenance Graph. Users can navigate relationships among knowledge sources, skills, tools, validation records, and historical versions, allowing them to understand how a skill was constructed, validated, reused, and evolved over time. This provenance-aware exploration mechanism improves the transparency and interpretability of large-scale skill ecosystems.

\subsection{Skill Governance}

Figure~\ref{fig:demo}(c) showcases the skill governance interface. SkillWiki manages skill modifications through a Git-style governance workflow rather than directly updating skill repositories. Governance activities are primarily performed by internal meta-skills and self-management agents, which autonomously generate maintenance proposals, analyze structured differences, conduct reviews, and manage skill releases. By default, the governance workflow can operate fully autonomously, enabling continuous maintenance and evolution of large-scale skill repositories. Meanwhile, users can inspect governance records, review proposed changes, and intervene at any stage when desired. User decisions always take precedence over autonomous governance actions. This design combines autonomous skill maintenance with human oversight, providing an auditable, traceable, and reversible mechanism for large-scale skill evolution.

\subsection{Skill Evolution}

Figure~\ref{fig:demo}(d) illustrates the skill evolution interface. SkillWiki continuously monitors skill execution through runtime feedback, including usage statistics, success rates, latency measurements, failure patterns, and reflection memories. Based on these signals, the system automatically evaluates skill health and identifies degraded, stale, or critical skills that require maintenance. When health issues or recurring failure patterns are detected, SkillWiki generates maintenance proposals for skill refinement, decomposition, replacement, or retirement, which are subsequently routed through the autonomous governance workflow. By connecting execution feedback, health assessment, and governance, SkillWiki enables skills to continuously adapt to changing environments and accumulated experience, transforming skill maintenance from a manual process into a self-sustaining evolution mechanism.

\section{Evaluation}

We evaluate SkillWiki from two perspectives. First, we study whether heterogeneous knowledge materials can be reliably transformed into governed skill assets. Second, we evaluate whether the resulting skills can be continuously maintained and evolved through the proposed governance and lifecycle framework.

\subsection{Knowledge-to-Skill Production}
We evaluate whether SkillWiki can transform heterogeneous knowledge materials into governed skill assets. The benchmark consists of 125 artifacts spanning five source types: trajectories, documents, API specifications, scripts, and historical skills (Appendix~\ref{appendix: knowledge artifact corpus}).

Table~\ref{tab:production} reports the results of the complete knowledge-to-skill workflow. Across 125 artifacts, 99 were successfully converted into governed skill candidates and integrated into the SkillWiki repository, demonstrating the feasibility of continuous skill production from diverse knowledge sources.

\begin{table}[t]
\centering
\small
\setlength{\tabcolsep}{5pt}
\begin{tabular}{lccc}
\toprule
\textbf{Source Type} &
\textbf{Artifacts} &
\textbf{Candidate} &
\textbf{Governed} \\
\midrule
Trajectory          & 25 & 24 & 24 \\
Document            & 25 & 17 & 17 \\
API Specification   & 25 & 24 & 24 \\
Script              & 25 & 18 & 18 \\
Historical Skills   & 25 & 16 & 16 \\
\midrule
\textbf{Total}      & \textbf{125} & \textbf{99} & \textbf{99} \\
\bottomrule
\end{tabular}
\caption{
Knowledge-to-skill production results.
}
\label{tab:production}
\end{table}

\subsection{Lifecycle Governance and Evolution}
To evaluate lifecycle governance and evolution, we conduct a full-chain case study using an API documentation artifact. The artifact progresses through skill production, verification, release, execution, repair, versioning, deprecation, and archival within SkillWiki.

Figure~\ref{fig:lifecycle} illustrates the resulting lifecycle trajectory. The skill traverses all lifecycle states (S0--S7), showing that SkillWiki supports end-to-end lifecycle management, including governance, evolution, version control, and retirement of skills.

\section{Conclusion}

We presented SkillWiki, a unified infrastructure for agent skill production, governance, and evolution. By connecting knowledge sources, skills, provenance, and execution feedback, SkillWiki enables skills to be continuously created, managed, and improved throughout their lifecycle. We hope SkillWiki provides a foundation for scalable and sustainable agent skill ecosystems.

\section*{Limitation}
SkillWiki focuses on the infrastructure for skill production, governance, and evolution. First, although the current system demonstrates end-to-end management of governed skill assets, its behavior in substantially larger repositories containing tens of thousands of skills has not yet been systematically evaluated. Second, our evaluation primarily measures the effectiveness of the proposed infrastructure and governance workflow, rather than the downstream impact of SkillWiki on diverse agent benchmarks and long-horizon tasks. Third, while SkillWiki supports execution-driven skill evolution, the long-term stability of continuously evolving skill ecosystems remains an open research question and requires further study under extended deployment settings.

\bibliography{custom}

\appendix

\section{Knowledge Artifact Corpus}
\label{appendix: knowledge artifact corpus}

\paragraph{Corpus Construction.}
To evaluate the knowledge-to-skill production capability of SkillWiki, we construct a corpus consisting of 125 heterogeneous knowledge artifacts. The corpus spans five source types: trajectories, documents, API specifications, scripts, and historical skills, with 25 artifacts collected for each category.

\paragraph{Data Sources.}
The corpus is curated from publicly available benchmarks, agent systems, software documentation, API specifications, and skill repositories. Representative sources include SkillsBench~\cite{li2026skillsbenchbenchmarkingagentskills}, WebArena~\cite{zhou2024webarenarealisticwebenvironment}, Anthropic Skills, and publicly available software documentation and workflow repositories. The resulting corpus covers diverse domains including software engineering, web automation, office productivity, data analysis, finance, security, and science.

Each artifact is managed through a manifest-driven workflow that records provenance information, source URLs, licensing metadata, source domains, and expected skill characteristics. This design enables reproducible evaluation while preserving traceability between generated skills and their originating knowledge sources.

\paragraph{Evaluation Protocol.}
During evaluation, each artifact is processed through the complete SkillWiki production pipeline, including knowledge parsing, candidate skill generation, provenance registration, graph construction, and lifecycle integration. Knowledge parsing and candidate generation are performed using DeepSeek-V4-Flash, while graph construction and lifecycle registration are handled by SkillWiki system services. The resulting candidates are incorporated into the SkillWiki governance framework and used to evaluate whether heterogeneous knowledge materials can be consistently transformed into governed skill assets.

\section{Runtime Monitoring and Health Signals}

SkillWiki continuously monitors released skills using runtime execution feedback.
The monitoring process collects four categories of signals:

(1) execution statistics, including execution counts and success rates;

(2) failure evidence and recurring failure patterns;

(3) reflection memories generated from execution failures and recovery attempts;

(4) lifecycle state transitions and maintenance proposals.

For a skill $s$, the success rate is defined as

\begin{equation}
SR(s)=
\frac{N_{\mathrm{success}}(s)}
     {N_{\mathrm{exec}}(s)},
\end{equation}

where $N_{\mathrm{exec}}(s)$ is the total number of executions and
$N_{\mathrm{success}}(s)$ is the number of successful executions.

The failure rate is defined as

\begin{equation}
FR(s)=
\frac{N_{\mathrm{failure}}(s)}
     {N_{\mathrm{exec}}(s)},
\end{equation}

where $N_{\mathrm{failure}}(s)$ denote the numbers of failed executions.

Reflection memories are accumulated from repeated execution failures and are used as evidence for maintenance proposal generation.

These signals are consumed by the monitoring and evolution modules to identify degraded skills and trigger maintenance workflows.

\end{document}